\documentclass{Interspeech2024}
\usepackage{algorithm}
\usepackage{algpseudocode}
\usepackage{url}
\usepackage{hyperref}
\usepackage[super]{nth}
\usepackage{subcaption}
\usepackage{multirow}
\usepackage{booktabs}
\usepackage{makecell}
\newcommand{\overbar}[1]{\mkern 1.5mu\overline{\mkern-1.5mu#1\mkern-1.5mu}\mkern 1.5mu}
\interspeechcameraready 
\title{Contrastive Feedback Mechanism for Simultaneous Speech Translation}

\name[affiliation={1}]{Haotian}{Tan}
\name[affiliation={1,2}]{Sakriani}{Sakti}
\address{
  $^1$Nara Institute of Science and Technology, Japan \\
  $^2$Japan Advanced Institute of Science and Technology, Japan 
  }
\email{tan.haotian.tf5@naist.ac.jp, ssakti@is.naist.jp}


\begin{document}

\maketitle

\begin{abstract}
Recent advances in simultaneous speech translation (SST) focus on the decision policies that enable the use of offline-trained ST models for simultaneous inference.
These decision policies not only control the quality-latency trade-off in SST but also mitigate the impact of unstable predictions on translation quality by delaying translation for more context or discarding these predictions through stable hypothesis detection.
However, these policies often overlook the potential benefits of utilizing unstable predictions. We introduce the contrastive feedback mechanism (CFM) for SST, a novel method that leverages these unstable predictions as feedback to improve translation quality. CFM guides the system to eliminate undesired model behaviors from these predictions through a contrastive objective.
The experiments on 3 state-of-the-art decision policies across 8 languages in the MuST-C v1.0 dataset show that CFM effectively improves the performance of SST.
\end{abstract}

\section{Introduction}
\label{section: 1}
% Problem statement
% The prediction based on limited acoustic information is unstable and harmful to the translation quality.
Simultaneous speech translation (SST) aims to mimic professional human interpreters to perform translation in real-time with low latency while maintaining high translation quality. In SST, unlike offline ST waiting till the end of the input sentence, it receives the incomplete speech inputs, namely speech segments or chunks, to execute translation incrementally.
However, some of the translations produced by the chunks are unstable since the earlier incremental steps often contain very limited context and the acoustic information towards the chunk endpoints tends to be uncertain \cite{lowlatency}.

% Literature review
Early works \cite{simulspeech, ma2020simulmt, zaidi2021decision, ustc} that trained SST systems with only partial input available have been shown to be dispensable \cite{papi2022does}. On the other hand, the methods of decision policy, which enable offline-trained ST models for simultaneous inference, have become increasingly popular over the past few years. Decision policies are rules to determine whether a system should emit predictions or wait for more input segments for better reliability.
The wait-k \cite{wait-k} is the most traditional decision policy, originally developed for text-to-text tasks and then adapted to the SST. It waits for k speech segments \cite{simulspeech} or words detected by the CTC prediction \cite{zeng2021realtrans} before the translation starts. Another type of decision policy is known as stable hypothesis detection. It detects and displays only the stable part of the predicted hypothesis and discards the remaining unstable predictions. The stable hypothesis detection enables more flexible wait/emit decision-making, thereby performing better than the wait-k. Among them, the Hold-n \cite{lowlatency} is the most straightforward policy, which takes the last n tokens as unstable predictions and deletes them. Furthermore, the local agreement (LA) policy was also proposed in \cite{lowlatency} and won the SST track of IWSLT 2022 \cite{cunikit,iwslt2022}. It detects stable and unstable predictions by making an agreement between two consecutive chunks. The SP-n policy \cite{cunikit,spn} is a variation of LA, which looks for the agreement in the beam of n consecutive chunks. However, its performance is worse than the vanilla LA policy \cite{cunikit}. Additionally, EDAtt \cite{EDATT} and AlignAtt \cite{alignatt} exploit the cross-attention mechanism to determine stable and unstable predictions and have demonstrated state-of-the-art (SOTA) performance in the quality-latency trade-off for SST.
\begin{figure}[t]
\centering
\begin{subfigure}[t]{\columnwidth}
\centering
\includegraphics[width=0.85\linewidth]{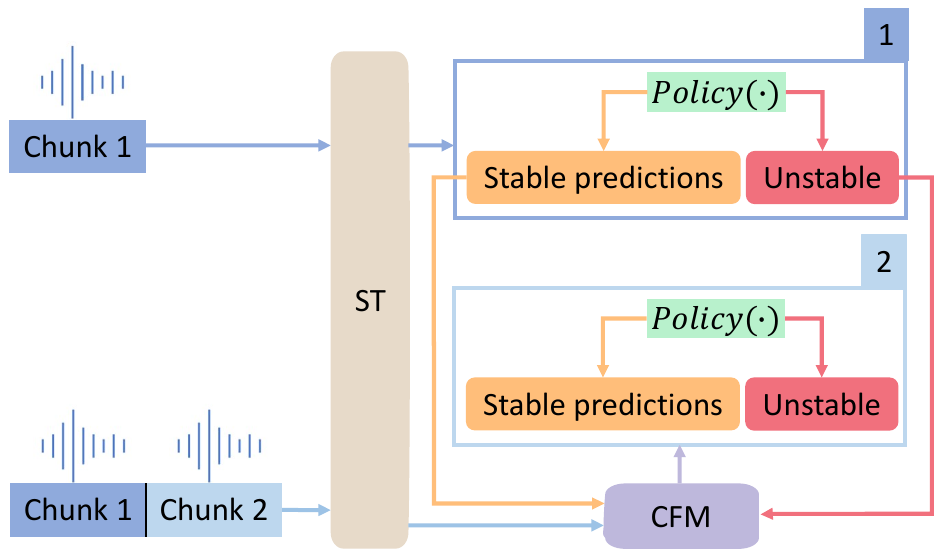}
% \caption{Framework of the SST with CFM.}
\label{fig:CFM_framework}
\end{subfigure}\hfill % maximize horizontal separation
\begin{subfigure}[t]{\columnwidth}
\centering
\includegraphics[width=0.95\linewidth]{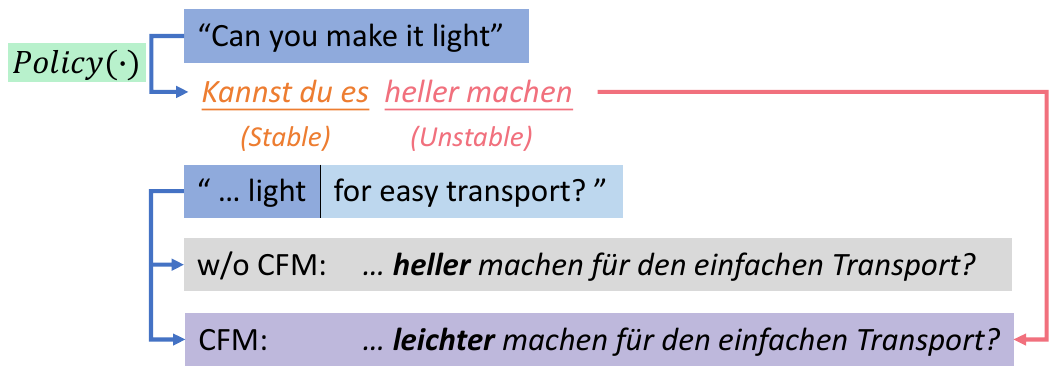}
% \caption{English-German translation example with/without CFM.}
\label{fig:CFM_example}
\end{subfigure}\hfill % maximize horizontal separation
\caption{Framework of the SST with CFM. \textbf{Top:} CFM leverages unstable predictions from an earlier chunk (marked as 1) as feedback to enhance the prediction of a subsequent chunk (marked as 2). \textbf{Bottom:} An English-German translation example with/without CFM. The word \textit{"light"} can be translated as \textit{"heller"} (illumination) or \textit{"leichter"} (weight). CFM helps to filter out the undesired model behavior of translating \textit{"light"} to \textit{"heller"} inappropriately.}
\label{fig:approach}
\end{figure}

Overall, these decision policies are made to reduce the unstable predictions by waiting for more context before translation starts (wait-k) or detecting and discarding them via stable hypothesis detection. In this paper, we explore employing unstable predictions to improve SST performance by proposing a contrastive feedback mechanism (CFM). Our proposal is based on the intuition that the unstable model predictions generated with less acoustic information have stronger tendencies to involve more undesired model behaviors (e.g., repetition, mistranslation) than those generated with a larger context. Specifically, we take the unstable predictions determined by the decision policies from an earlier chunk as feedback information. The CFM filters out the undesired model behaviors highlighted by the feedback to enhance the translation quality of a later chunk. The proposed CFM is expected to generate translations that emphasize the best of the larger context predictions and remove the undesired tendencies of the unstable prediction feedback. To the best of our knowledge, this is the first approach that improves SST performance by exploiting unstable predictions via a feedback mechanism without extra training or modifying the decision policies. In general, our contributions are as follows:
\begin{itemize}
\setlength\itemsep{-0.1em}
    \item We propose CFM, a novel method for SST that leverages unstable model predictions of earlier chunks to improve the SST performance of later chunks.
    \item We introduce chunk size as an alternative principle factor that affects the quality-latency trade-off of AlignAtt and EDAtt policies.
    \item We demonstrate that the unstable predictions generated by limited context can be utilized to improve the translation quality of SST.
\end{itemize}

\section{Contrastive Feedback Mechanism}
\label{section:approach}
As illustrated in Figure \ref{fig:approach}, the decision policy determines the stable predictions for user display, while the rest are considered unstable. Unlike conventional SST systems that ignore the use of unstable predictions, the proposed CFM utilizes these predictions generated by the earlier chunk as feedback to assist the model generation of the later chunk. For instance, consider the English-German translation example in Figure \ref{fig:approach}. The ST model is inclined to translate the word \textit{"light"} as \textit{"heller"} (illumilation) given the first chunk context, \textit{"Can you make it light"}. Even with the extended context \textit{"Can you make it light for easy transport"} provided in a later chunk, the model can still assign a large probability mass to the word \textit{"heller"}. CFM can filter out this undesired model behavior to obtain a more appropriate translation, \textit{"leichter"} (weight).

\subsection{CFM-Enhanced SST}
\label{section: 2.1}
We incorporate the CFM into SST with beam search and describe the chunk-level process in Algorithm 1.
\begin{algorithm}
\caption{Contrastive Feedback Mechanism for SST}
\begin{algorithmic}
    \Require $speech\_segment, emitted\_tokens, P_{f}$
    \State $beams \gets \text{InitializeBeams}(emitted\_tokens)$
    \If {$decoding\_step==0$ and $P_{f}$ is not None}
    \State $ P_{c} \gets \text{Predict}(speech\_segment
    ,beams) $
    \State $beams \gets beams + \text{CFM-Score}(P_{c},P_{f})$
    \Else
    \State $best\_beam \gets \text{Predict}(speech\_segment, beams)$
    \State $StablePredictions, NewP_{f} \gets \text{Policy}(best\_beam)$
    \State $P_{f} \gets NewP_{f}$
    \State $emitted\_tokens \gets StablePredictions$
    \EndIf
\end{algorithmic}
\end{algorithm}
Specifically, we obtain the feedback information $P_{f}$ from unstable predictions of an earlier chunk based on their probability distributions, and we utilize this feedback information to rescore the candidate tokens at the initial decoding step of current chunk through the CFM-Score:
\begin{equation}
    \text {CFM-Score}\left(P_{c} ; P_{f}\right)=\log{P_{c}} + \text {Contrast}\left(P_{c} ; P_{f}\right) 
\end{equation}
where $P_{c}$ indicates the initial probability distribution of the current chunk. In this way, the candidate tokens are rescored by a contrastive objective, $\text {Contrast}\left(P_{c} ; P_{f}\right)$, rather than relying solely on log-probabilities. 
Inspired by \cite{contrastive}, which contrasts a larger language model (LM) with a smaller LM, we present the contrastive objective for SST to measure the difference between likelihood under a larger context input and a smaller context input:
\begin{equation}
\begin{aligned}
& \text {Contrast}\left(P_{c} ; P_{f}\right) 
& = \begin{cases}\log \frac{p_{c}\left(y_i \mid y_{<i}\right)}{P_{f}}, & \text { if } y_i \in \mathcal{V}_{\beta} \\
- \text { inf, } & \text { otherwise}\end{cases}
\end{aligned}
\end{equation}
where token $y_i$ denotes the first token of the current chunk. Additionally, the plausibility constraint $\mathcal{V}_{\beta}$ is introduced to avoid potential issues with the contrastive objective:
\begin{equation}
\resizebox{0.89\hsize}{!}{
    $\begin{aligned}
        & \mathcal{V}_{\beta}=\\
        & \left\{y_i \in \mathcal{V}: p_{c}\left(y_i \mid y_{<i}\right) \geq \beta \max _\omega p_{c}\left(\omega \mid y_{<i}\right)\right\}
    \end{aligned}$}
\end{equation}
where $\mathcal{V}$ is the vocabulary list, and we set the constraint factor $\beta=0.1$ throughout the paper to truncate the probability distribution of token $y_i$.
For example, when $p_{c}\left(y_i \mid y_{<i}\right)=1\times10^{-3}$ and $P_{f}=1\times10^{-9}$, although a very low probability score of $1\times10^{-3}$ is given by the larger context prediction, a very high reward of $\text{Contrast}=6.0$ is assigned. In this case, the contrastive objective would amplify some undesired model behaviors of the larger context prediction.
Under this constraint, the tokens assigned low probabilities by the larger context prediction are ignored. The contrastive objective emphasizes the best of the larger context prediction $P_c$ and removes the undesired tendencies of the feedback $P_f$ by providing a reward or penalty to each candidate token in $P_{c}$ according to their differences to the feedback information, $P_{f}$. In general, CFM is expected to factor out the undesired model behaviors hiding in the feedback information to improve the initial prediction quality of the current chunk. This, in turn, enhances the overall chunk-level translation quality through the autoregressive decoder.

\subsection{Feedback Information}
\label{section: 2.2}
CFM can serve all decision policies that follow the stable hypothesis detection manner as we introduced in Section \ref{section: 1} to improve the SST performance. In this paper, we demonstrate its effectiveness based on three state-of-the-art decision policies. We use different ways to select the feedback information, $P_{f}$, that may better reflect the undesired model behaviors to achieve the goal.
\begin{itemize}
    \item \textbf{AlignAtt} \cite{alignatt}: The state-of-the-art decision policy for SST detects unstable predictions based on the cross-attention scores. If the output prediction $y_i$ mostly attends to the last $f$ frames, where $f$ is a hyperparameter that controls the quality-latency trade-off, the emission stops and the predictions $y_{\geq{i}}$ are taken as unstable predictions. We take the averaged probability distribution of the unstable predictions $\overbar{P}_{\geq{i}}$ as the feedback information for the AlignAtt policy.
    \item \textbf{EDAtt} \cite{EDATT}: The first decision policy for SST employs the attention mechanism to guide the inference. For the prediction $y_i$, if the sum of its last $\lambda$ frames' attention scores is larger than a threshold $\alpha$, the emission stops. The $\lambda$ is an empirically defined hyperparameter, and the $\alpha$ controls the quality-latency trade-off. Similar to AlignAtt, we take $\overbar{P}_{\geq{i}}$ as the feedback information.
    \item \textbf{Local Agreement (LA)} \cite{lowlatency}: Different from the attention-based AlignAtt and EDAtt that can make decisions chunk-by-chunk, the LA policy needs two chunks to make a decision. It compares the predictions of two consecutive chunks and only emits their longest common prefixes (i.e., agreement). The remaining predictions of the later chunk are seen as unstable and will be used to compare with the next chunk predictions. The speech segment length (aka. chunk size) controls the quality-latency trade-off for LA. In this paper, we take the probability distribution of the first unstable prediction as feedback information for the LA policy.
\end{itemize}

\section{Experimental Setup}
\subsection{Data}
We evaluate the effectiveness of the proposed method in eight languages of the widely used MuST-C v1.0\footnote{Data publicly available at: \url{https://mt.fbk.eu/must-c/}} tst-COMMON dataset \cite{must}: English (en)$\rightarrow$ \{German (de), Spanish (es), French (fr), Italian (it), Dutch (nl), Portuguese (pt), Romanian (ro), Russian (ru)\}.

\subsection{Offline ST Model}
The offline performance of the ST model provides a topline for its simultaneous manner by consuming the full context to generate translation. To guarantee the translation qualities for all target languages, we follow \cite{alignatt} to build the offline ST model, which consists of a Conformer \cite{conformer} encoder and a Transformer \cite{attention} decoder and achieves SOTA performance on the MuST-C v1.0 dataset. The implementation is based on the Fairseq toolkit \cite{wang2020fairseq, ott2019fairseq}.
\subsection{Decision Policies}
\label{section: 3.3}
% trade-off control, parameter setting
We evaluate our proposed method based on three decision policies, AlignAtt, EDAtt, and LA, as we introduced in Section \ref{section: 2.2}. Following the authors' setting, we extract the attention weights from the \nth{4} decoder layer for the attention-based AlignAtt and EDAtt policies and set $\lambda=2$ for the EDAtt policy. We vary $f$ in [2, 4, 6, 8, 10, 12, 14] for AlignAtt and $\alpha$ in [0.6, 0.4, 0.2, 0.1, 0.05, 0.03] for EDAtt to control their quality-latency trade-off, respectively. For the LA policy, we use different chunk sizes of [0.4s, 0.6s, 0.8s, 1.0s, 1.2s] to control its quality-latency trade-off. Additionally, we also experiment with chunk sizes of 0.2s, 0.4s, 1s, and 2s to investigate how the chunk size affects the quality-latency trade-off for AlignAtt and EDAtt policies. We show later in Section \ref{section: chunk} that the chunk size is a significant factor that affects their simultaneous performance. In this paper, we use chunk size=1s for both AlignAtt and EDAtt policies.
\subsection{Evaluation}
% Metrics; SimulEVAL, Device
The SacreBLEU ($\uparrow$) \cite{sacrebleu} is utilized to measure translation quality. Moreover, we use the SimulEval Toolkit \cite{simuleval} for simultaneous evaluation and measure the translation latency using the computational-aware version of LAAL ($\downarrow$) \cite{laal}. All experiments are conducted on the NVIDIA A40 GPU with 44GB of RAM.
\section{Experiments and Results}
\subsection{Offline Results}
\begin{figure}[ht]
\centering
\includegraphics[width=0.85\columnwidth]{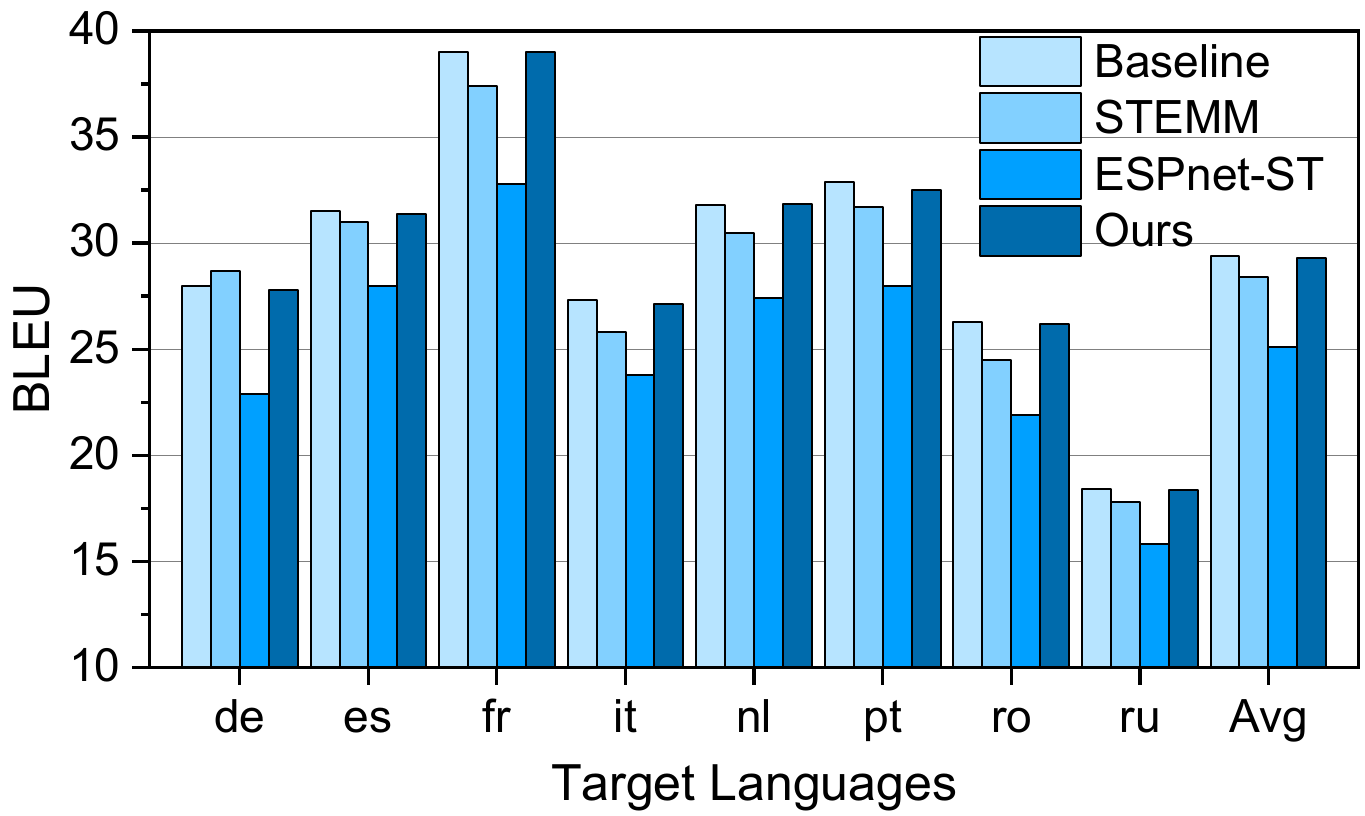}
\caption{Offline translation quality comparison of different ST models.}
\label{fig: offline}
\end{figure}

We compare our reproduced offline ST model with the baseline model \cite{alignatt}, STEMM \cite{stemm}, and ESPnet-ST \cite{espnet}, which are also trained on the MuST-C v1.0 dataset. As illustrated in Figure \ref{fig: offline}, our reproduced offline model achieves better translation quality than the STEMM except for the en$\rightarrow$de direction, and it significantly outperforms the ESPnet-ST on all translation directions. The offline ST model we employ for simultaneous translation demonstrates performance comparable to that of the state-of-the-art baseline model. Its robustness ensures that the translations produced simultaneously are also coherent and meaningful.
\subsection{Chunk Size}
\label{section: chunk}
\begin{figure}[htbp]
\begin{minipage}[t]{\columnwidth}
\centering
  \includegraphics[width=0.5\columnwidth]{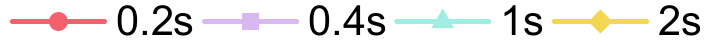}
\end{minipage}
\begin{subfigure}[t]{0.5\columnwidth}
    \centering
    \includegraphics[width=0.95\linewidth]{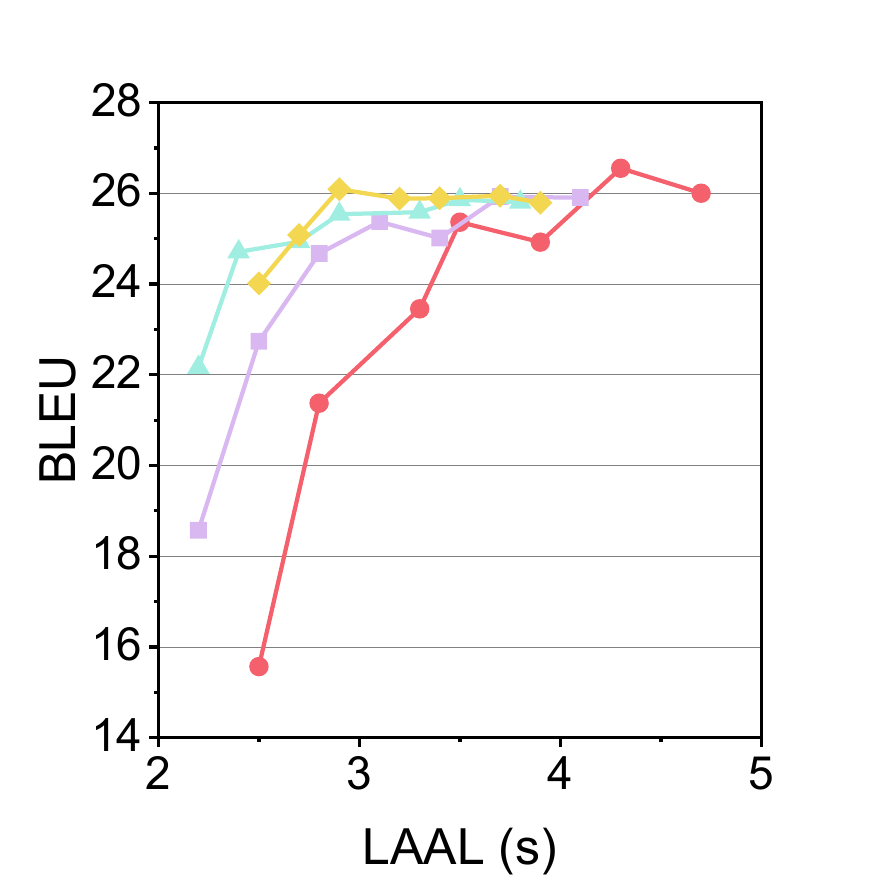}
    \caption{AlignAtt}
    \label{fig:alignatt_chunk}
\end{subfigure}\hfill % maximize horizontal separation
\begin{subfigure}[t]{0.5\columnwidth}
    \centering
    \includegraphics[width=0.95\linewidth]{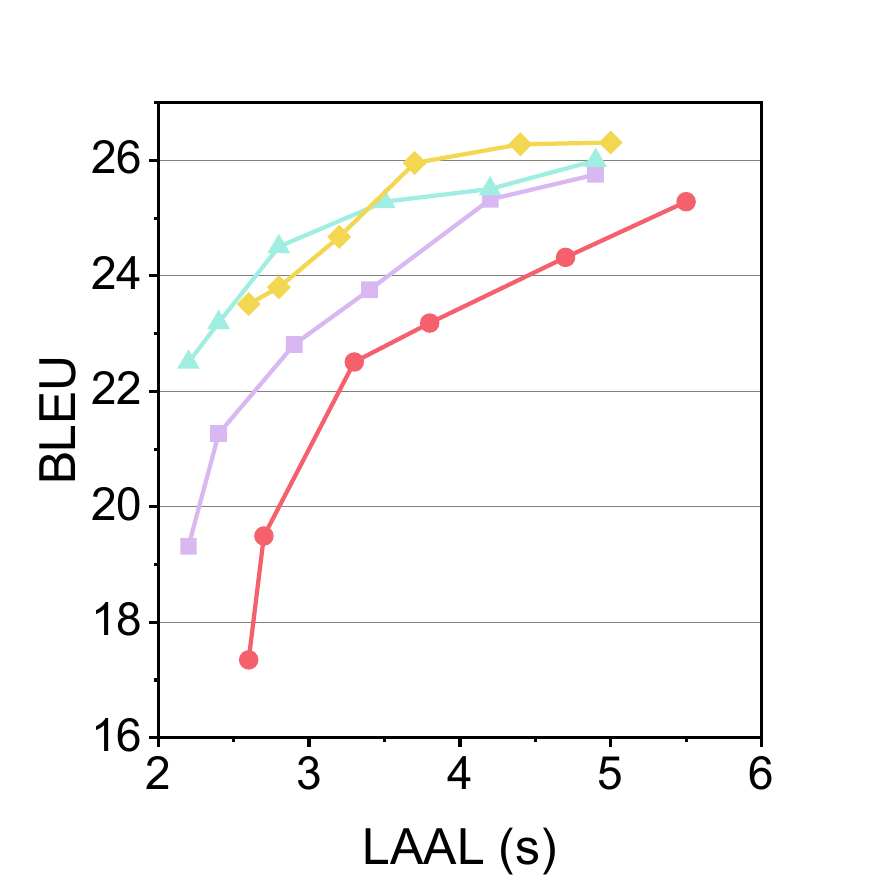}
    \caption{EDAtt}
    \label{fig:edatt_chunk}
\end{subfigure}\hfill
\caption{Quality-latency trade-off of different chunk sizes combined with AlignAtt and EDAtt policies.}
\label{fig: chunk}
\end{figure}

We experiment with the influence of using different chunk sizes for the AlignAtt and EDAtt policies. The results in Figure \ref{fig: chunk} document that the chunk size has a significant impact on the performance for both AlignAtt and EDAtt policies. Especially in the lower latency region,
\begin{table*}[ht!]
\centering
\scalebox{0.64}{
\begin{tabular}{c|cccccccccccccccc}
\cline{2-17}
                          & \multicolumn{4}{c|}{de}                                                   & \multicolumn{4}{c|}{es}                                                   & \multicolumn{4}{c|}{fr}                                                   & \multicolumn{4}{c}{it}                                 \\ \cline{2-17} 
                          & \multicolumn{2}{c|}{w/o CFM} & \multicolumn{2}{c|}{CFM}                   & \multicolumn{2}{c|}{w/o CFM} & \multicolumn{2}{c|}{CFM}                   & \multicolumn{2}{c|}{w/o CFM} & \multicolumn{2}{c|}{CFM}                   & \multicolumn{2}{c|}{w/o CFM} & \multicolumn{2}{c}{CFM} \\ \cline{2-17} 
                          & BLEU          & LAAL         & BLEU           & \multicolumn{1}{c|}{LAAL} & BLEU          & LAAL         & BLEU           & \multicolumn{1}{c|}{LAAL} & BLEU          & LAAL         & BLEU           & \multicolumn{1}{c|}{LAAL} & BLEU          & LAAL         & BLEU            & LAAL  \\ \hline
\multirow{7}{*}{AlignAtt} & 22.64         & 1.93         & 22.73          & 1.93                      & 25.85         & 1.91         & 25.86          & 1.91                      & 32.80         & 1.96         & 32.81          & 1.96                      & 21.61         & 1.89         & 21.63           & 1.90  \\
                          & 24.90         & 2.20         & 24.93          & 2.18                      & 28.08         & 2.20         & 28.12          & 2.18                      & 35.53         & 2.20         & 35.54          & 2.23                      & 23.88         & 2.15         & 23.93           & 2.16  \\
                          & 26.07         & 2.46         & 26.09          & 2.44                      & 29.30         & 2.43         & 29.42          & 2.46                      & 36.87         & 2.47         & 36.89          & 2.47                      & 25.16         & 2.42         & 25.22           & 2.42  \\
                          & 26.63         & 2.70         & 26.63          & 2.69                      & 30.03         & 2.68         & 30.11          & 2.70                      & 37.64         & 2.73         & 37.68          & 2.74                      & 25.97         & 2.70         & 26.05           & 2.70  \\
                          & 26.91         & 2.96         & \textbf{27.01} & 2.98                      & 30.53         & 2.93         & 30.61          & 2.95                      & 37.93         & 3.01         & 37.94          & 2.98                      & 26.22         & 2.94         & 26.29           & 2.96  \\
                          & 27.25         & 3.20         & 27.28          & 3.21                      & 30.56         & 3.17         & 30.71          & 3.19                      & 38.27         & 3.21         & \textbf{38.33} & 3.21                      & 26.54         & 3.16         & 26.69           & 3.16  \\
                          & 27.33         & 3.41         & 27.36          & 3.41                      & 30.79         & 3.40         & \textbf{30.95} & 3.43                      & 38.40         & 3.42         & 38.42          & 3.44                      & 26.56         & 3.40         & \textbf{26.75}  & 3.40  \\ \hline
\multirow{6}{*}{EDAtt}    & 23.71         & 2.03         & 23.73          & 2.07                      & 27.05         & 1.99         & 27.18          & 2.04                      & 33.89         & 2.07         & 33.92          & 2.11                      & 23.02         & 1.99         & 23.14           & 2.04  \\
                          & 24.77         & 2.24         & 24.82          & 2.27                      & 27.92         & 2.14         & 28.00          & 2.17                      & 35.04         & 2.24         & 35.12          & 2.27                      & 23.89         & 2.14         & 23.95           & 2.19  \\
                          & 26.01         & 2.70         & \textbf{26.10} & 2.72                      & 29.29         & 2.38         & 29.42          & 2.44                      & 36.55         & 2.58         & 36.68          & 2.60                      & 24.67         & 2.42         & \textbf{24.91}  & 2.45  \\
                          & 26.74         & 3.18         & 26.78          & 3.23                      & 30.08         & 2.69         & 30.27          & 2.74                      & 37.45         & 2.99         & \textbf{37.59} & 3.02                      & 25.71         & 2.76         & 25.84           & 2.79  \\
                          & 27.08         & 3.80         & 27.10          & 3.81                      & 30.60         & 3.07         & 30.77          & 3.13                      & 38.01         & 3.43         & 38.10          & 3.46                      & 26.50         & 3.20         & 26.53           & 3.20  \\
                          & 27.35         & 4.26         & 27.38          & 4.34                      & 30.82         & 3.44         & \textbf{31.19} & 3.47                      & 38.50         & 3.84         & 38.53          & 3.86                      & 26.69         & 3.54         & 26.86           & 3.57  \\ \hline
\multirow{5}{*}{LA}       & 20.64         & 2.30         & \textbf{22.00} & 2.48                      & 23.09         & 2.24         & \textbf{24.88} & 2.46                      & 29.42         & 2.20         & \textbf{31.47} & 2.35                      & 19.80         & 2.43         & \textbf{21.30}  & 2.51  \\
                          & 24.23         & 2.52         & 24.67          & 2.68                      & 27.37         & 2.46         & 27.70          & 2.60                      & 34.69         & 2.47         & 34.88          & 2.59                      & 23.23         & 2.62         & 24.22           & 2.70  \\
                          & 25.98         & 2.76         & 26.08          & 2.88                      & 29.15         & 2.69         & 29.40          & 2.83                      & 36.92         & 2.72         & 36.98          & 2.83                      & 25.13         & 2.85         & 25.56           & 2.94  \\
                          & 27.02         & 2.99         & 27.12          & 3.12                      & 30.05         & 2.93         & 30.37          & 3.07                      & 37.79         & 2.93         & 37.82          & 3.03                      & 26.06         & 3.03         & 26.14           & 3.13  \\
                          & 27.22         & 3.25         & 27.24          & 3.34                      & 30.37         & 3.17         & 30.76          & 3.29                      & 38.29         & 3.18         & 38.31          & 3.22                      & 26.49         & 3.24         & 26.75           & 3.34  \\ \hline
\end{tabular}
}
\caption{Simultaneous results of AlignAtt, EDAtt, and LA policies with/without CFM for en$\rightarrow$\{de, es, fr, it\} translations. The bold values denote the maximum improved translation quality achieved by different policies for each language after applying CFM.}
\label{table: 1}
\end{table*}
\begin{table*}[ht!]
\centering
\scalebox{0.64}{
\begin{tabular}{c|cccccccccccccccc}
\cline{2-17}
                          & \multicolumn{4}{c|}{nl}                                                   & \multicolumn{4}{c|}{pt}                                                   & \multicolumn{4}{c|}{ro}                                                   & \multicolumn{4}{c}{ru}                                 \\ \cline{2-17} 
                          & \multicolumn{2}{c|}{w/o CFM} & \multicolumn{2}{c|}{CFM}                   & \multicolumn{2}{c|}{w/o CFM} & \multicolumn{2}{c|}{CFM}                   & \multicolumn{2}{c|}{w/o CFM} & \multicolumn{2}{c|}{CFM}                   & \multicolumn{2}{c|}{w/o CFM} & \multicolumn{2}{c}{CFM} \\ \cline{2-17} 
                          & BLEU          & LAAL         & BLEU           & \multicolumn{1}{c|}{LAAL} & BLEU          & LAAL         & BLEU           & \multicolumn{1}{c|}{LAAL} & BLEU          & LAAL         & BLEU           & \multicolumn{1}{c|}{LAAL} & BLEU          & LAAL         & BLEU            & LAAL  \\ \hline
\multirow{7}{*}{AlignAtt} & 25.89         & 1.83         & 25.96          & 1.86                      & 27.28         & 1.98         & 27.32          & 1.98                      & 22.86         & 1.95         & 22.88          & 1.94                      & 14.45         & 1.93         & 14.47           & 1.93  \\
                          & 28.31         & 2.09         & 28.40          & 2.12                      & 29.52         & 2.22         & 29.58          & 2.24                      & 24.66         & 2.17         & 24.70          & 2.20                      & 15.64         & 2.17         & 15.73           & 2.17  \\
                          & 29.43         & 2.37         & 29.56          & 2.36                      & 31.05         & 2.51         & 31.08          & 2.51                      & 25.51         & 2.46         & 25.57          & 2.48                      & 16.67         & 2.42         & 16.79           & 2.42  \\
                          & 29.97         & 2.60         & 30.07          & 2.61                      & 31.22         & 2.76         & 31.37          & 2.76                      & 25.91         & 2.76         & 25.93          & 2.75                      & 17.25         & 2.68         & 17.36           & 2.69  \\
                          & 30.49         & 2.85         & \textbf{30.68} & 2.87                      & 31.68         & 3.01         & \textbf{31.97} & 3.01                      & 26.23         & 2.99         & 26.31          & 3.01                      & 17.58         & 2.95         & \textbf{17.75}  & 2.96  \\
                          & 30.87         & 3.08         & 31.04          & 3.11                      & 32.02         & 3.24         & 32.18          & 3.27                      & 26.12         & 3.22         & 26.26          & 3.28                      & 17.80         & 3.18         & 17.91           & 3.19  \\
                          & 31.10         & 3.32         & 31.19          & 3.32                      & 32.19         & 3.50         & 32.23          & 3.48                      & 26.27         & 3.45         & \textbf{26.44} & 3.47                      & 17.98         & 3.44         & 18.14           & 3.46  \\ \hline
\multirow{6}{*}{EDAtt}    & 27.21         & 1.92         & 27.28          & 1.96                      & 28.60         & 2.09         & 28.94          & 2.14                      & 23.99         & 2.01         & 24.12          & 2.09                      & 15.34         & 2.03         & 15.46           & 2.09  \\
                          & 28.19         & 2.10         & 28.24          & 2.15                      & 29.51         & 2.22         & 30.04          & 2.30                      & 24.85         & 2.15         & 24.96          & 2.23                      & 16.05         & 2.26         & \textbf{16.29}  & 2.31  \\
                          & 29.44         & 2.46         & 29.56          & 2.50                      & 30.24         & 2.52         & \textbf{30.77} & 2.58                      & 25.73         & 2.43         & \textbf{25.88} & 2.48                      & 17.14         & 2.78         & 17.31           & 2.82  \\
                          & 30.38         & 2.89         & 30.42          & 2.92                      & 31.26         & 2.85         & 31.55          & 2.90                      & 25.97         & 2.76         & 26.11          & 2.79                      & 17.84         & 3.48         & 18.03           & 3.49  \\
                          & 30.91         & 3.40         & \textbf{31.05} & 3.45                      & 31.86         & 3.33         & 31.94          & 3.37                      & 26.26         & 3.20         & 26.32          & 3.23                      & 18.28         & 4.52         & 18.50           & 4.53  \\
                          & 31.32         & 3.84         & 31.41          & 3.87                      & 32.08         & 3.73         & 32.11          & 3.77                      & 26.37         & 3.61         & 26.44          & 3.68                      & 18.30         & 5.40         & 18.37           & 5.42  \\ \hline
\multirow{5}{*}{LA}       & 22.19         & 2.17         & \textbf{24.43} & 2.40                      & 24.30         & 2.32         & \textbf{26.14} & 2.51                      & 20.17         & 2.30         & \textbf{21.60} & 2.50                      & 12.98         & 2.37         & \textbf{14.32}  & 2.68  \\
                          & 26.98         & 2.45         & 27.94          & 2.60                      & 28.50         & 2.53         & 29.24          & 2.74                      & 23.63         & 2.53         & 24.29          & 2.68                      & 15.52         & 2.61         & 16.25           & 2.85  \\
                          & 29.25         & 2.71         & 29.69          & 2.82                      & 30.46         & 2.77         & 30.80          & 2.92                      & 25.07         & 2.77         & 25.29          & 2.91                      & 16.95         & 2.87         & 17.30           & 3.05  \\
                          & 29.98         & 2.96         & 30.24          & 3.10                      & 31.41         & 3.03         & 31.48          & 3.16                      & 25.90         & 3.00         & 26.05          & 3.12                      & 17.29         & 3.12         & 17.76           & 3.29  \\
                          & 30.57         & 3.18         & 30.69          & 3.27                      & 31.87         & 3.25         & 31.92          & 3.36                      & 26.27         & 3.25         & 26.38          & 3.34                      & 17.88         & 3.37         & 17.95           & 3.49  \\ \hline
\end{tabular}
}
\caption{Simultaneous results of AlignAtt, EDAtt, and LA policies with/without CFM for en$\rightarrow$\{nl, pt, ro, ru\} translations. The bold values denote the maximum improved translation quality achieved by different policies for each language after applying CFM.}
\label{table: 2}
\end{table*}
the performance gaps between using a larger chunk size of 2s and a smaller chunk size of 0.2s achieve 8.5 and 6.2 BLEU points for AlignAtt and EDAtt, respectively. Although increasing the chunk size improves the translation quality, we set the chunk size to 1s for both AlignAtt and EDAtt policies in this paper since this setting achieves the lowest latency while maintaining comparable translation quality with a 2s chunk size. The experimental results demonstrate the significance of appropriately setting the chunk size for AlignAtt and EDAtt policies before practical use.
\begin{figure}[ht]
\centering
\includegraphics[width=0.85\columnwidth]{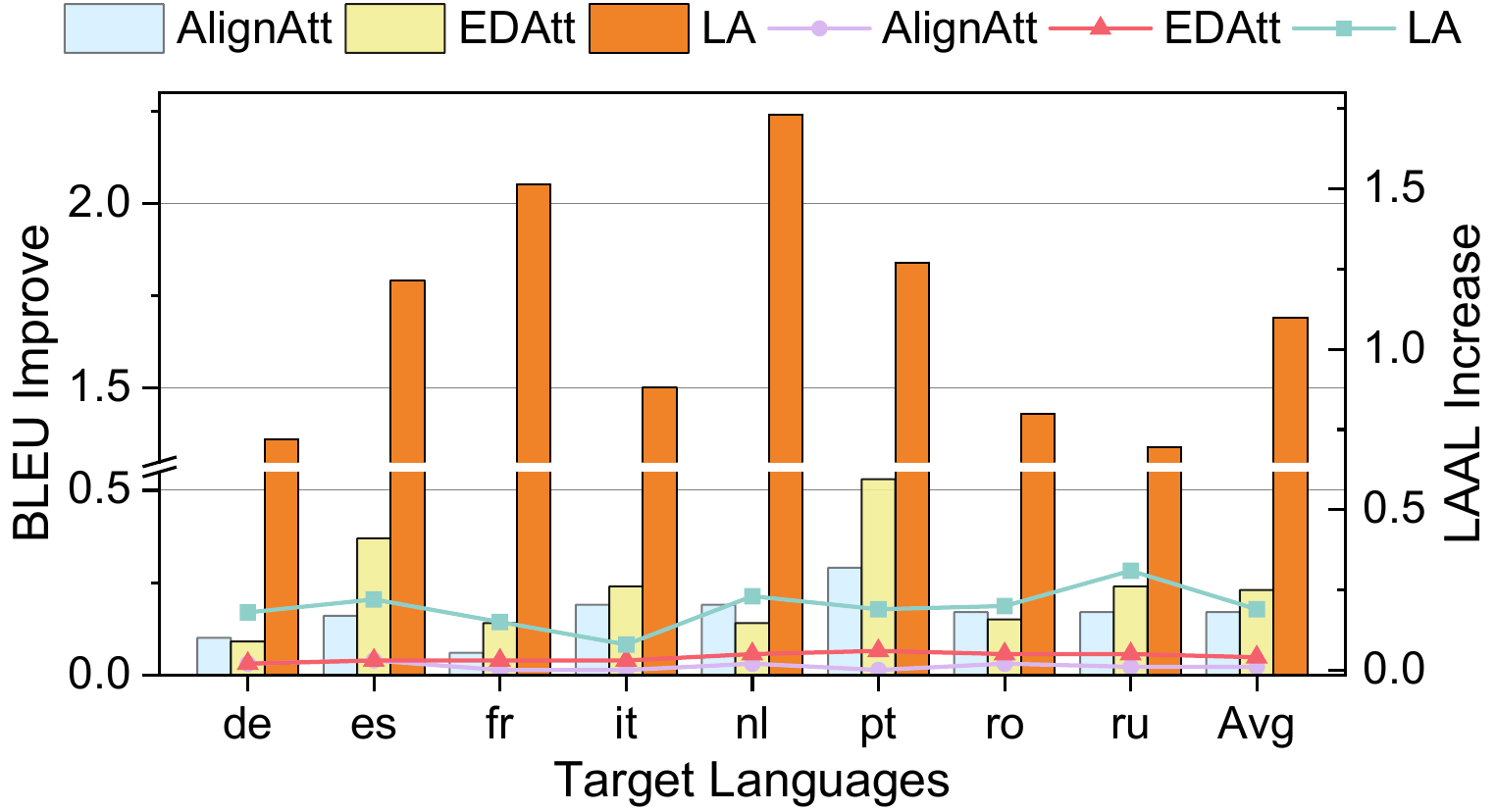}
\caption{Maximum BLEU score improvements (shown in column charts) and their associated latency increases (depicted in line charts) correspond to bold values in Tables \ref{table: 1} and \ref{table: 2}.}
\label{fig: max_improve}
\end{figure}

\subsection{Simultaneous performance of CFM}
We present numerical results\footnote{Corresponding to different quality-latency controls in Section \ref{section: 3.3}.} for translation quality (BLEU $\uparrow$) and latency (LAAL $\downarrow$) with or without the proposed CFM across eight translation directions in Tables \ref{table: 1} and \ref{table: 2}. As we can see, applying CFM effectively improves translation quality for three state-of-the-art decision policies with negligible latency increases across eight translation directions.
Figure \ref{fig: max_improve} details the maximum improvements achieved with CFM across eight languages.
For the AlignAtt policy, CFM yields a maximum improvement of 0.29 BLEU points, with a maximum latency increase of only 0.03s. The EDAtt policy sees a maximum improvement of 0.53 BLEU points, with a maximum latency increase of 0.05s. Notably, CFM is most effective on the LA policy, with significant translation quality improvements spanning from 1.34 to 2.05 BLEU points, accompanied by a latency increase of up to 0.22s. 
For the most notable improvement of 2.05 BLEU points achieved on en$\rightarrow$nl translation with LA policy, the vanilla LA achieves 22.19 BLEU points with a 95\% confidence interval (CI)\footnote{We follow the default setting of \cite{Confidence_Intervals} to evaluate 95\% CI.} of [21.54, 22.85], while the CFM-enhanced LA achieves 24.43 BLEU points, 95\% CI [23.74, 25.09].

Overall, CFM improves translation quality across all decision policies with negligible additional latency, showcasing its effectiveness in enhancing SST performance.

\section{Conclusion}
We proposed CFM, a novel method for SST that exploits unstable predictions from an earlier chunk to enhance the translation quality of a later chunk through a feedback mechanism.
CFM improves the translation quality based on the incremental process of SST rather than doing extra training or modifying the decision policies.
Evaluations on three state-of-the-art decision policies across eight languages have demonstrated the effectiveness of CFM, achieving a significant maximum improvement of up to 2.05 BLEU points with a negligible sacrifice in latency.

\section{Acknowledgements}
Part of this work is supported by JSPS KAKENHI Grant Numbers JP21H05054 and JP23K21681.
\bibliographystyle{IEEEtran}
\bibliography{mybib}

\end{document}